\documentclass[sn-basic,iicol]{sn-jnl}
\usepackage{bbding}
\usepackage{algorithm}
\usepackage{algorithmicx}
\usepackage{algpseudocode}
\usepackage{graphicx}
\usepackage{subfigure}
\usepackage{color}

\usepackage{xcolor}

\jyear{2022}
\theoremstyle{thmstyleone}%

\theoremstyle{thmstyletwo}%
\theoremstyle{thmstylethree}%

\begin{document}

\title{A Multi-view Mask Contrastive Learning Graph Convolutional Neural Network for Age Estimation}

\author[1]{\fnm{Yiping} \sur{Zhang}}\email{yipingzhang@csuft.edu.cn}

\author[1]{\fnm{Yuntao} \sur{Shou}}\email{yuntaoshou@csuft.edu.cn}

\author*[1]{\fnm{Tao} \sur{Meng}}\email{mengtao@hnu.edu.cn}

\author[1]{\fnm{Wei} \sur{Ai}}\email{aiwei@hnu.edu.cn}

\author[2]{\fnm{Keqin} \sur{Li}}\email{lik@newpaltz.edu}

\affil[1]{\orgdiv{School of Computer and Information Engineering}, \orgname{Central South University of Forestry and Technology}, \orgaddress{ \city{Changsha}, \postcode{410082}, \state{Hunan}, \country{China}}}
\affil[2]{\orgdiv{Department of Computer Science}, \orgname{State University of New York}, \orgaddress{\state{New Paltz}, \city{New York 12561},  \country{USA}}}

\abstract{The age estimation task aims to use facial features to predict the age of people and is widely used in public security, marketing, identification, and other fields. However, the features are mainly concentrated in facial keypoints, and existing CNN and Transformer-based methods have inflexibility and redundancy for modeling complex irregular structures. Therefore, this paper proposes a Multi-view Mask Contrastive Learning Graph Convolutional Neural Network (MMCL-GCN) for age estimation. Specifically, the overall structure of the MMCL-GCN network contains a feature extraction stage and an age estimation stage. In the feature extraction stage, we introduce a graph structure to construct face images as input and then design a Multi-view Mask Contrastive Learning (MMCL) mechanism to learn complex structural and semantic information about face images. The learning mechanism employs an asymmetric siamese network architecture, which utilizes an online encoder-decoder structure to reconstruct the missing information from the original graph and utilizes the target encoder to learn latent representations for contrastive learning. Furthermore, to promote the two learning mechanisms better compatible and complementary, we adopt two augmentation strategies and optimize the joint losses. In the age estimation stage, we design a Multi-layer Extreme Learning Machine (ML-IELM) with identity mapping to fully use the features extracted by the online encoder. Then, a classifier and a regressor were constructed based on ML-IELM, which were used to identify the age grouping interval and accurately estimate the final age. Extensive experiments show that MMCL-GCN can effectively reduce the error of age estimation on benchmark datasets such as Adience, MORPH-II, and LAP-2016.}


\maketitle

\vspace{5pt}

\section{Introduction}\label{sec1}
Age estimation plays a significant role in human-computer interaction and computer vision, with promising applications in face retrieval, social media, market analysis, human-computer interaction, and security monitoring \cite{yang2018ssr, shou2022conversational, shou2022object, ying2021prediction, meng2024multi, shou2023low}. However, various intrinsic factors such as gender and race and extrinsic factors such as facial expressions, facial decoration, dressing, body postures, and background environmental factors make age estimation complex and challenging \cite{agbo2021deep, rothe2015dex, shou2023graph, shou2023czl, shou2023comprehensive, ai2023gcn}.

In the past few years, there has been a flurry of research in this area, with various approaches emerging \cite{chang2010ranking, duan2017ensemble, meng2023deep, shou2023adversarial, SHOU2024102590, shou2023masked, shou2024revisiting}. Typically, traditional age estimation approaches could be categorized into five different types: classification methods, regression methods, label distribution, ranking methods, and hybrid methods \cite{agbo2021deep, duan2020egroupnet, shou2024efficient, ai2023two, meng2024revisiting}. Traditional methods are mainly based on handcrafted models that employ image processing filtering techniques for feature extraction and machine learning algorithms, such as decision trees and support vector machines. However, these techniques are insufficient to cope with the complexity of real-world environments and practical applications, as they cannot effectively deal with facial images that exhibit variations in scale, rotation, and lighting \cite{duan2017ensemble, ricanek2006morph, ai2024gcn}.

To address the challenges posed by these complex environments, current age estimation methods leverage convolutional neural networks (CNNs) to recognize challenging facial images and achieve robust age estimation \cite{li2019bridgenet, rothe2015dex, gurpinar2016kernel}. Li \emph{et al.} \cite{li2019bridgenet} proposed using local regressors to delineate age data regions, gated network simulations to construct regression forests to provide continuous age weights, and regression-weighted combinations to obtain estimated ages. Rothe \emph{et al.} \cite{rothe2015dex} proposed age prediction based on the expectation values of convolutional networks for classification while building an IMDB-WIKI face dataset as a pre-training dataset. Gürpinar \emph{et al.} \cite{gurpinar2016kernel} proposed a two-stage system for estimating with a face detector based on a deformable partial CNN model, and a kernel extreme learning machine is used for classification. Zhang \emph{et al.} \cite{zhang2019c3ae} presented C3AE leveraging a multi-scale feature stitching approach to compensate for the small model fitting problem through two levels of serial supervision.

In addition, with the great fire of the Transformer in the vision field, Xue \emph{et al.} \cite{xue2021transfer} proposed TransFER to learn different relation-aware local representations of facial expression features based on the transformer. Wan \emph{et al.} \cite{wan2022facial} proposed a novel spatial FAT for Face Attribute Transformer (FAT) and its variants to achieve high-quality makeup migration. FAT can model the deep semantic correspondence and interaction between source and reference faces and accurately estimate and convey face attributes. Moreover, Xia \emph{et al.}\cite{xia2022sparse} proposed the Sparse Local Patch Transformer (SLPT) method to learn the intrinsic relationship between key feature points, which can better capture facial information, and its effect exceeds the previous CNN model.

\begin{figure*}
	\centering
	\includegraphics[width=1\textwidth]{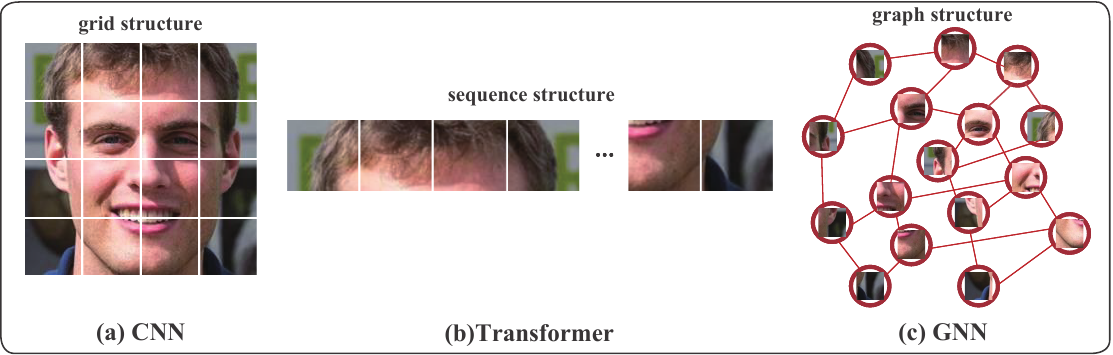}
	\caption{The illustration of CNN, Transformer, and GNN on image representation. a) CNN uses convolution operators to extract features on images with grid structure. b) Transformer uses the attention mechanism to extract features on images with sequence structure. c) GNN uses the information aggregation mechanism to extract features on images with graph structure. The graph structure encompass sequences and grids as varied instances capturing long-range contextual information. Therefore, GNN extracts image features with more flexibility and little redundancy.} 
	\label{Fig.1}
\end{figure*}

Nevertheless, although CNNs and Transformers have achieved good results in many fields, they process images, text, and other data using the traditional Euclidean space, i.e., structured data for representation, which is inflexible and redundant for image data with complex structural semantics. Especially for human faces, the main information is concentrated around the key points of the human faces, including the mouth, eyebrows, eyes, nose, etc. As shown in Fig. \ref{Fig.1}, the emergence of Graph Neural Networks (GNNs) \cite{han2022vision} provides an effective method for complex structural semantic feature extraction. In order to minimize the excessive proliferation of nodes, ViG (Vision gnn) draws on ViT's concept of segmentation to divide the image into smaller patches, a division that introduces complex dependencies between nodes and their nearest neighbors to construct an adaptive graph. Compared with CNNs and Transformers, utilizing GNNs \cite{zhang2023gcg} can overcome the limitations of limited receptive fields and regular locations, thus effectively capturing long-range contextual information in non-Euclidean spaces. Therefore, we believe that GNNs can better deal with the key features of faces and their connections.

To better implement vision tasks, using GNN alone is not enough to realize its potential \cite{hou2022graphmae, huang2023masked, zhang2023gcg}. For example, when faced with limited face datasets, we may need to combine self-supervised learning (SSL). There are two families of self-supervised approaches: contrastive SSL and generative SSL. Contrastive learning (CL) is the dominant self-supervised learning paradigm in the pre-training stage by employing simple discriminations to extract features, which bring the distance between positive samples closer and pushes the distance between negative samples further. Relying on robust data augmentation, MoCo \cite{he2020momentum} and SimCLR \cite{chen2020simple} are two iconic contrastive learning methods that use a siamse network structure to maximize the consistency of learned representations between similar samples. Corresponding to contrastive learning, masked image modeling (MIM) as generative SSL, enlightened by the idea of masked language modeling (MLM) in NLP \cite{devlin2019bert}, randomly masks several image patches and then reconstructs the masked image patches in an autoencoding manner. BeiT \cite{bao2021beit} used a pre-trained discrete variational autoencoder (dVAE) for image token sorting. MAE \cite{he2022masked} proposed an asymmetric encoder-decoder framework that reconstructs the original pixels from the underlying representation. Moreover, at the graph level \cite{hou2022graphmae}, Variational graph auto-encoders (VGAE) predicted the edge of loss, embedding propagation (EP) proposed to restore the vertex features, and GPT-GNN proposed an autoregressive architecture for iteratively performing node and edge reconstruction. However, contrastive learning and masked image modeling both have deficiencies and limitations. Masked image modeling optimizes local features to achieve the reconstruction goal but lacks semantic information about the learned features. In contrast, contrastive learning learns latent discriminability through relationships between different images, leading to more semantically meaningful representations. Therefore, considering the merits and defects of both methods, we integrate the two methods better to extract different distinctive features for the input image.

Motivated by the analysis above, we propose a Multi-view Mask Contrastive Learning Graph Convolutional Neural Network (MMCL-GCN) for age estimation. MMCL-GCN is a hybrid structure. First, a graph convolutional neural network (GCN) is designed to extract image features, and then a classifier and regressor based on an extreme learning machine (ELM) are designed to estimate age. Specifically, the overall structure of the MMCL-GCN network contains a feature extraction stage and an age estimation stage. In the feature extraction stage, we introduce a graph structure to construct face images as input and then design a multi-view mask contrastive learning mechanism to learn complex structural and semantic information about face images. The learning mechanism employs an asymmetric siamese network architecture, which utilizes an online encoder-decoder structure to reconstruct the missing information from the original graph and utilizes the target encoder to learn latent representations for contrastive learning. Furthermore, to promote the two learning mechanisms better compatible and complementary, we adopt the primary node masking
strategy for the online encoder, the secondary edge dropping strategy for the target encoder and optimize the joint losses. Both encoder and decoder can use GNN as the backbone network, which can be GCN \cite{kipf2017semi}, GIN \cite{velivckovic2017graph}, or GAT \cite{xu2019powerful}. In the age estimation stage, to fully use the image features extracted by the online encoder, we choose the extreme learning machine instead of the traditional softmax method for the classification and regression tasks of age estimation. Technically, to deal with high-dimensional features more effectively and avoid model degradation, we design a multi-layer extreme learning machine (ML-IELM) with identity mapping. Then, a classifier and a regressor were constructed based on ML-IELM, which were used to identify the age grouping interval and accurately estimate the final age. Our contributions to this paper are summarized as follows:

\begin{itemize}
	\item
	A novel image feature extraction method with a Multi-view Mask Contrastive Learning framework, i.e., MMCL, is proposed. MMCL can effectively fuse contrastive learning and mask image reconstruction to learn latent discriminative features and high-level local features of images.
	
	\item
	A multi-layer extreme learning machine with identity mapping (ML-IELM) is designed, and constructing classifiers and regressors based on ML-IELM can effectively reduce the age estimation error.

    \item
	Finally, extensive experiments on three popular benchmark datasets, MORPH-II, LAP-2016 and Adience, show that MMCL-GCN outperforms existing comparison algorithms in terms of mean absolute error and normal score ($\epsilon$-error).
\end{itemize}

\section{Related Work}\label{sec2}
In this section, the self-supervised methods are first introduced, including masked image autoencoders and contrastive learning. Then we briefly describe the frontier research of graph neural networks. Finally, we retrospect several related works in age estimation, especially those deep learning-based methods.

\subsection{Contrastive Learning}
Contrastive learning has received significant attention in various self-supervised and unsupervised paradigms \cite{chen2020simple, he2020momentum, le2020contrastive}. It entails learning feature representations from unlabeled data and using them in downstream tasks. In the learning process, it constructs positive and negative instances such that positive instances are closer together in the projection space, and negative instances are pushed away.

Contrastive Multi-view Coding (CMC) \cite{le2020contrastive}, an early adopter of multi-view contrastive learning, demonstrated the flexibility of contrastive learning and the feasibility of multi-view multimodality. After that, MoCo \cite{he2020momentum} and SimCLR \cite{chen2020simple} are two important foundational works that push the boundaries of contrastive learning. MoCo is very similar to Inst Disc in that it utilizes a queue instead of a memory bank to store negative samples, allowing it to take more negative examples from contrastive learning. Moreover, MoCo replaced the constraint term in the loss with a momentum encoder, thus achieving a better result with a momentum update encoder. In SimCLR, the authors pointed out that strong data augmentation is crucial for contrastive learning, and they also made use of a negative sample during the training period. Most importantly, they suggested using a non-linear projection head and maximizing the protocols expressed in the plan. BYOL drew on the core principles of MoCo and SimCLR and introduced an online encoder after the projection head to predict the output of the momentum encoder. Furthermore, SimSiam \cite{chen2021exploring} proposed the simple Siamese networks by replacing momentum updates through stop-gradient techniques. Based on siamese networks \cite{chen2021exploring} and extensions of MoCo and BYOL, MoCo-v3 \cite{chen2021empirical} and DINO \cite{caron2021emerging} leveraged visual transformers (ViT) as the backbone of their models with better success.

Although contrastive learning has achieved good performance in the self-supervised domain of images, it focuses on learning global discriminative features and lacks the attention of local information.

\subsection{Masked Image Encoding}
Mask coding is a standard method for generative self-supervised learning for pre-training, which aims to reconstruct the missing information of the original data, and was originally widely used in the field of natural language processing (NLP) \cite{devlin2019bert, bao2021beit}.

In NLP, Masked Language Modelling (MLM) \cite{devlin2019bert} has shown great versatility in a variety of downstream tasks. BEiT \cite{bao2021beit} is pioneering work in transferring the BERT form of pre-training to the field of vision and introducing the task concept of MIM pre-training. Aligned with the success of ViT \cite{dosovitskiy2021image}, masked modeling extends to visual tasks with great potential. MAE (Masked Autoencoders) \cite{he2022masked} proposed an alternative ViT backbone network that uses random masks to process the input image blocks, and direct reconstruction of masked image blocks for training. SimMIM \cite{xie2022simmim} proposes a simple mask learning framework, MaskFeat, which achieves SOTA results. In MaskFeat \cite{wei2022masked}, the reconstruction target changed from the original pixel values to the low-level local features (HOG) descriptors, which no longer relied on large datasets and improved pre-training efficiency.

Although masked image modeling has achieved good performance, autoencoders encourage preserving information in the latent representation rather than modeling the relationship between different images for contrastive learning. Several current works have proposed methods that use a similar siamese structure assembling two components \cite{assran2022masked, huang2023contrastive}. This motivates us to further investigate combining contrastive learning and masking image modeling at the graph level to develop an enhanced pre-training model that learns different data features in a complementary manner.

\subsection{Graph Neural Networks}
Graph neural networks (GNNs) \cite{han2022vision} have become an increasingly popular method for studying graph-structured data. GNNs achieve state-of-the-art performance in various tasks such as node classification, link prediction, and graph classification by aggregating information iteratively through neighboring nodes and updating the node representations \cite{xu2019powerful}.

Early GNNs mainly focused on how to aggregate node information in a graph structure. Micheli \emph{et al.} \cite{micheli2009neural} proposed an early formation of spatially based graph convolutional networks through structurally compound non-recursive layers. Bruna \emph{et al.} \cite{bruna2013spectral} first proposed graph-based convolution on spectral graph theory. Hamilton \emph{et al.} \cite{hamilton2017inductive} proposed GraphSAGE for efficient computation and generalization capabilities by means of parameter sharing. Casanova \emph{et al.} \cite{velivckovic2017graph} developed the graph attention network (GAT), which computed the hidden representation of each node and followed a strategy of self-attentive clustering of neighbors. Liu \emph{et al.} \cite{vashishth2019composition} proposed the Generative Interaction Network (GIN) to learn the iterative message-passing procedures on graphs. Jiang \emph{et al.}  \cite{jiang2021gecns} proposed local graph elastic convolutional networks (GeCNs) for sparse graph data representation learning.

Moreover, as self-supervised methods have made a splash in computer vision, graph neural networks have also begun to explore these methods. Petar \emph{et al.} \cite{velickovic2019deep} proposed DGI (Deep Graph Infomax) using contrastive learning on graph-structured data. Hou \emph{et al.} \cite{hou2022graphmae} proposed GraphMAE utilizing masked graph autoencoders, which could outperform graph contrastive learning models. Shahid \emph{et al.} \cite{shahid2023view} proposed VA-GNN utilizing an unsupervised reposition approach and jointly designed a VA neural network to achieve skeleton-based recognition of human actions. Xu  \emph{et al.} \cite{xu2022topology} proposed Topology-aware CNN (Ta-CNN), which implement the topology modeling power of GCNs using a CNN. Ju \emph{et al.} \cite{Ju2023multi} introduced PARETOGNN achieving self-supervised by manifold pretext tasks observing multiple philosophies.

With the continuous promotion of the application range of GNNs, Kai Han \emph{et al.} \cite{han2022vision} recently proposed the GNN-based visual task processing architecture ViG and achieved state-of-the-art performance in target detection. Since ViG can process image data without being limited by the location of features and can capture potentially complex structural relationships between features, more and more researchers use ViG as the backbone network for visual tasks. For example, Yu Zhang \emph{et al.} \cite{zhang2023gcg} proposed a GCG-Net network for geographic image classification in combination with a grouping optimization strategy. Junjun Huang \emph{et al.} \cite{huang2023masked} proposed an MGNN network for real-time polyp detection using a masking strategy.

The above methods are well-explored for applying GNNs to processing vision tasks. This paper explores a new direction of graph representation learning for vision tasks, combining generative SSL and contrastive SSL to achieve more powerful vision performance.

\subsection{Age Estimation}
In computer vision, advances in deep learning and countless studies have achieved better results in age estimation \cite{agbo2021deep}. These studies can be categorized as single or hybrid structure-based approaches \cite{duan2018hybrid}.

In a single-structure-based age estimation study, Chen \cite{chang2010ranking} proposed a ranking-CNN for age prediction that has a range of basic networks trained with regular labels. Hassncer \emph{et al.} \cite{levi2015age} proposed a shallow CNN structure containing three convolutional layers and two fully designed layers to learn feature representations. Rothe \emph{et al.} \cite{rothe2015dex} estimate appearance age based on VGG-16 and the DEX (deep expectation of apparent age), the framework is pre-trained on ImageNet, and the regression task is defined as a classification task with softmax expectation refinement.

Furthermore, fusing several techniques has become popular in age estimation due to hybrid architectures' complementary advantages. Duan \emph{et al.} \cite{duan2018hybrid} proposed an ensemble structure called CNN2ELM, which consists of a three-layer model, including feature extraction, feature fusion, and age prediction. First, they trained three networks corresponding to attributes such as age, gender, and race to extract and fuse different features from images of the same person. Then, based on the extreme learning machine, they constructed an age interval classifier and a specific age regressor for final age prediction. Then Li \emph{et al.} \cite{li2019bridgenet} proposed BridgeNet for age estimation, consisting of two components: a gated network and a local regressor which could learn together in an end-to-end manner.

However, the above age estimation methods are all based on CNNs, which can only model the grid structure of images. In addition, single-layer extreme learning machines cannot handle high-dimensional data well. In this paper, we study age estimation based on hybrid structures, exploring the combination of GNN and ELM models to enhance age estimation performance and achieve as compact as possible while being more robust.

\section{METHODOLOGY}
\begin{figure*}
	\centering
	\includegraphics[width=1\textwidth]{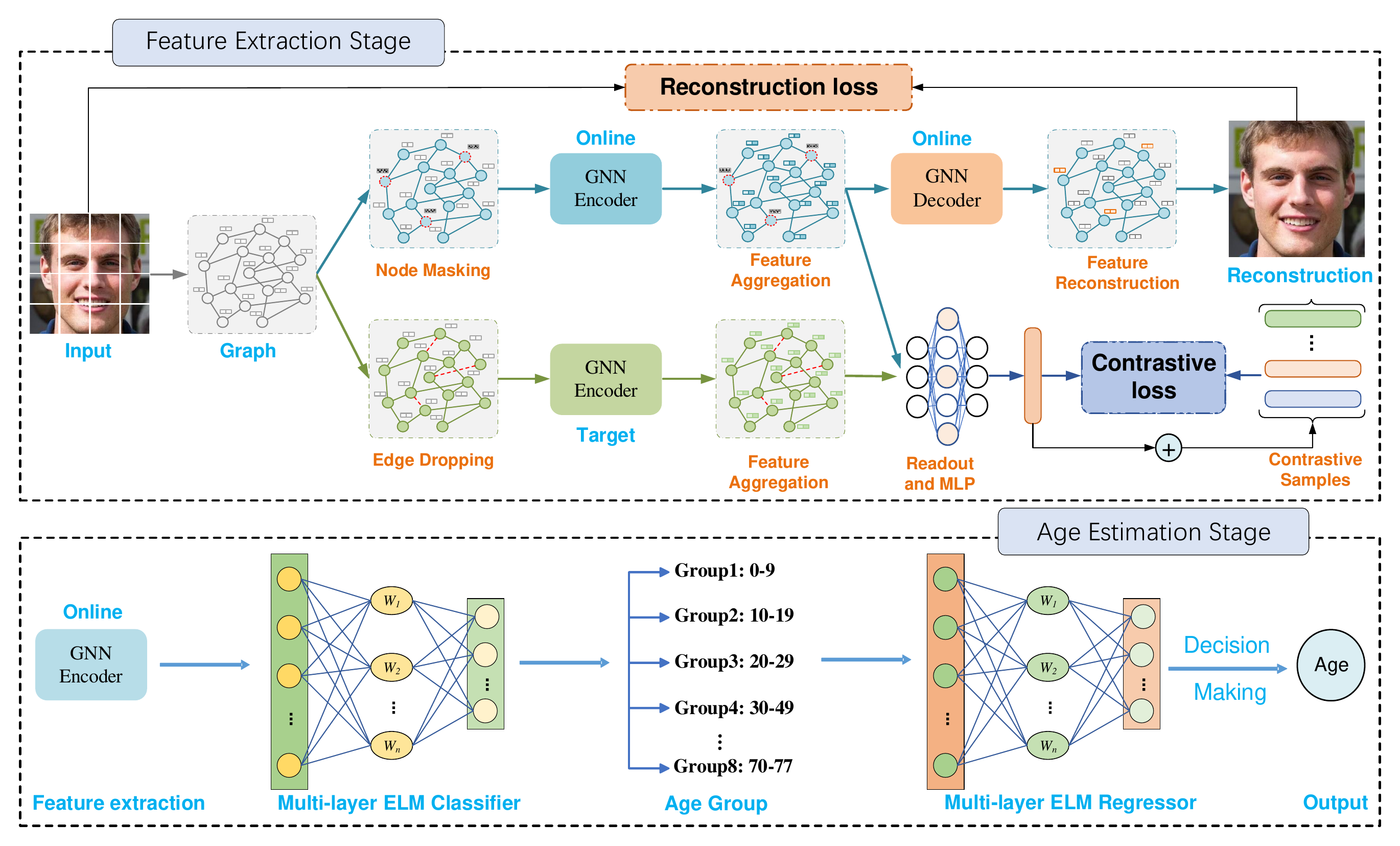}
	\caption{The overall framework of MMCL-GCN contains two stages: feature extraction stage and age estimation stage. In the feature extraction stage, we introduce a graph structure to construct face images as input and then design a Multi-view Mask Contrastive Learning (MMCL) mechanism to learn complex structural and semantic information about face images. MMCL utilizes an online encoder-decoder structure to reconstruct the masked patches from the original graph and a target encoder for contrastive learning. Moreover, it adopts the graph augmentation strategies and optimizes the joint losses to promote better compatible and complementary.} In the age estimation stage, to achieve robust age estimation, we use a multi-layer ELM classifier and regressor with identity mapping for preliminary age grouping and final age prediction, respectively.
	\label{Fig.2}
\end{figure*}

This section details the proposed approach of multi-view mask contrastive learning (MMCL) for image feature extraction and multi-layer extreme learning machine with identity mapping (ML-IELM) for age estimation. The overall framework is shown in Fig. \ref{Fig.2}.

\subsection{Overview}
The complete process of MMCL-GCN consists of two stages: the feature extraction stage and the age estimation stage. Before giving a specific design, we encountered and answered the following questions.

\textbf{Q1:} \emph{How to better exploit the potential of graph neural networks in vision tasks?}

\textbf{Q2:} \emph{Is it feasible to combine contrastive learning and masked image modeling and how to design an effective combination?}

\textbf{Q3:} \emph{How to design architecture for more accurate and robust age prediction?}

In the feature extraction stage, we introduce constructing images via graph structures as input and design multi-view mask contrastive learning (MMCL) to learn complex structural semantics for vision tasks. \textbf{Q1:} Our learning mechanism is an effective combination of contrastive learning, which can learn more discriminative global features, and mask reconstruction, which can learn important local features for restoration. \textbf{Q2:} Moreover, MMCL utilizes an online encoder-decoder structure to reconstruct the masked information from the original graph and a target encoder for contrastive learning. After MMCL optimizes a reconstruction loss and a contrastive objective, we obtain a robust and efficient online encoder for graph-level feature extraction. In the age estimation stage, we use the facial features extracted by the online encoder for age estimation. \textbf{Q3:} To achieve more accurate and robust age estimation, we designed a multi-layer extreme learning machine (ML-IELM) with identity mapping. Then we utilized an ML-IELM classifier for preliminary age grouping and an ML-IELM regressor for final age prediction.

\subsection{Framework of MMCL}
Multi-view mask contrastive learning mainly consists of graph construction and graph feature extraction. Below we describe these two components in detail.

\textbf{Graph Construction}: For an input 2D image $I$, whose resolution is $\left(H, W\right)$ and the channel is $C$, $I\in \mathbb{R} ^{H\times W\times C}$ is converted and tokenized to $N$ patches, where $P^2$ is the resolution of each image patch, and the resulting number of patches is $N = HW/P^2$. Then all patches are embedded into $M$ feature dimension, the feature vector $X=\left\{ x_ {i=1}^{n} \right\} \in \mathbb{R} ^{N\times M}$. Moreover, we obtain a graph $\mathcal{G}\left(V, E\right)$ where V denotes all the nodes, $E$ denotes all the edges, $D$ denotes the degree matrix and the connectivity of the nodes is computed by the adjacent matrix $A= \mathbb{R} ^{n\times n}$ ($n$ is the number of nodes). Specifically, let the feature node $v_i$  has $K$ nearest neighbors $\Bbbk \left(v_i \right)$ and indicate the edge from node $i$ to node $j$ by $e_{ij}$.

\textbf{Graph Feature Extraction}: After constructing a graph based on image features, we investigate the methods to extract graph features through GNNs. When transforms the image patches as graph node feature vectors, GNNs process them for efficient information exchange between image blocks, and compute the feature distances to determine the nearest neighbor. This graph representation learning has significant advantages in that it not only extracts richer information through a larger receptive field but also overcomes the deficiency of neural networks in processing images through a regular grid, which facilitates the detection of irregular and complex defects.

A common expression for a graph convolutional neural network (GCN) is:
\begin{equation}
	\begin{aligned}
		H^{l+1}=\sigma \left( \tilde{D}^{- \frac{1}{2}}\tilde{A}\tilde{D}^{-\frac{1}{2}}H^l\varTheta ^l \right)
	\end{aligned}
	\label{equ1}
\end{equation}
where $ \tilde{A}=A+I $, $I$ is the unit matrix, $\tilde{D}=\sum_ j{ \tilde{A}_{ij}}$ is the degree matrix, $H^0$ of the input layer is $X$, $H^l$ is the embedding of the graph nodes in the $l$-th layer, $\varTheta ^l$ is the trainable weight and is the activation function.

GCNs basically follows a strategy of aggregating information about its neighbors, i.e., by aggregating the representations of neighboring nodes and then updating the representation of the current node. The structural information of the $k$-th neighborhood could be captured after $k$ iterations, and a $k$-layer graph neural network generalization can be expressed as:
\begin{equation}
	\begin{aligned}
		a_v^{(k)} = AG{G^{(k)}}\left( {\left\{ {h_{{v^\prime }}^{(k - 1)}:{v^\prime } \in \Bbbk(v)} \right\}} \right)
	\end{aligned}
	\label{equ2}
\end{equation}

\begin{equation}
	\begin{aligned}
		h_{v}^{\left( k \right)}=COM^{\left( k \right)}\left( h_{v}^{\left( k-1 \right)},a_{v}^{k} \right)
	\end{aligned}
	\label{equ3}
\end{equation} where $a_{v}^{\left( k \right)}$ is the feature embedding of aggregated neighbor nodes, $h_{v}^{\left( k \right)}$ is the $k$-th layer embedding of the vertex, $\Bbbk \left(v_i \right)$ denotes a set of vertices adjacent to $ v_i$, and $AGG^{\left( k \right)}\left( \cdot \right)$ and  $COM^{\left( k \right)}\left( \cdot \right)$ are the aggregation function and activation function of the GNN layer. In different cases, it is necessary to choose different $AGG\left(  \cdot  \right)$ function and $COM\left(  \cdot  \right)$ function. In GCN, average pooling is used instead of maximum pooling, and the $AGG\left(  \cdot  \right)$ function and $COM\left(  \cdot  \right)$ function is defined as:
\begin{equation}
	\begin{split}
		h_v^{(k)} = ReLU\left( {W \cdot MEAN\left( {\left\{ {h_{{v^\prime }}^{(k - 1)}:{v^\prime } \in {\Bbbk k}(v)} \right\}} \right)} \right)
	\end{split}
	\label{equ4}
\end{equation}
After $k-$layer propagation, the output embedding of $\mathcal{G}$ is summarized at the layer by $READOUT\left( \cdot \right) $ function. Finally, the node features are aggregated to obtain graph-level features $h_\mathcal{G}$ and adopt a multi-layer perceptron (MLP) for downstream task:
\begin{equation}
	\begin{aligned}
		h_{\mathcal{G}}=R E A D O U T\left(\left\{h_v^{(k)} \mid v \epsilon \mathcal{G}\right\}\right)
	\end{aligned}
	\label{equ5}
\end{equation}
\begin{equation}
	\begin{aligned}
		z_{\mathcal{G}}=MLP\left( h_{\mathcal{G}} \right)
	\end{aligned}
	\label{equ6}
\end{equation}

\textbf{Multi-view Mask Contrastive Learning}: In particular, our architecture adopts an asymmetric siamese network that has an online encoder, an online decoder, and a target encoder. The online encoder-decoder learns local latent features by reconstructing the masked information, and the target encoder provides the online encoder contrastive supervision to learn the strong instance discriminability.

Let $\mathcal{F} _o\left( \cdot \right)$ as the online encoder, $\mathcal{F} _t\left( \cdot \right)$ as the target encoder, $\mathcal{H} _o\left( \cdot \right)$ as the online decoder, the parameters of $\mathcal{F} _o\left( \cdot \right)$ as $\varTheta _o$ and $\mathcal{F} _t\left( \cdot \right)$ as $\varTheta _t$. The encoder  maps the feature vector  $X=\left\{ x_{i=1}^{n} \right\}$ to embedding feature $H=\left\{ h_ {i=1}^{n} \right\}$. Generally, reconstruct the input image as:
\begin{equation}
	\begin{aligned}
		H=\mathcal{F} \left( A,X \right) =\sigma \left( \tilde{D}^{-\frac{1}{2}}\tilde{A}\tilde{D}^{-\frac{1}{2}}H^l\varTheta ^l \right)
	\end{aligned}
	\label{equ7}
\end{equation}
\begin{equation}
	\begin{aligned}
		\mathcal{G} ^{\prime}=\mathcal{H} \left( A,H \right)
	\end{aligned}
	\label{equ8}
\end{equation}

The type of online encoder and target encoder are the same expressive GNNs while the online decoder is a shallow GNN. In the optimization process, to slow down the parameter changes of the target network, different from the online encoder updates parameters $\varTheta _o$ by back propagation, the target encoder updates parameters $\varTheta _t$ by exponential moving average \cite{he2020momentum}:
\begin{equation}
	\begin{aligned}
		\varTheta _t\gets m\varTheta _t+\left( 1-m \right) \varTheta _o
	\end{aligned}
	\label{equ9}
\end{equation}
where $m\in \left[0,1\right)$ is the momentum hyperparameter and we set to 0.999. The momentum update rule propagates the update gradually from $\varTheta _t$ to $\varTheta _o$ , so that $\varTheta _o$ evolves smoothly and consistently.

\begin{figure}
	\centering
	\includegraphics[width=0.48\textwidth]{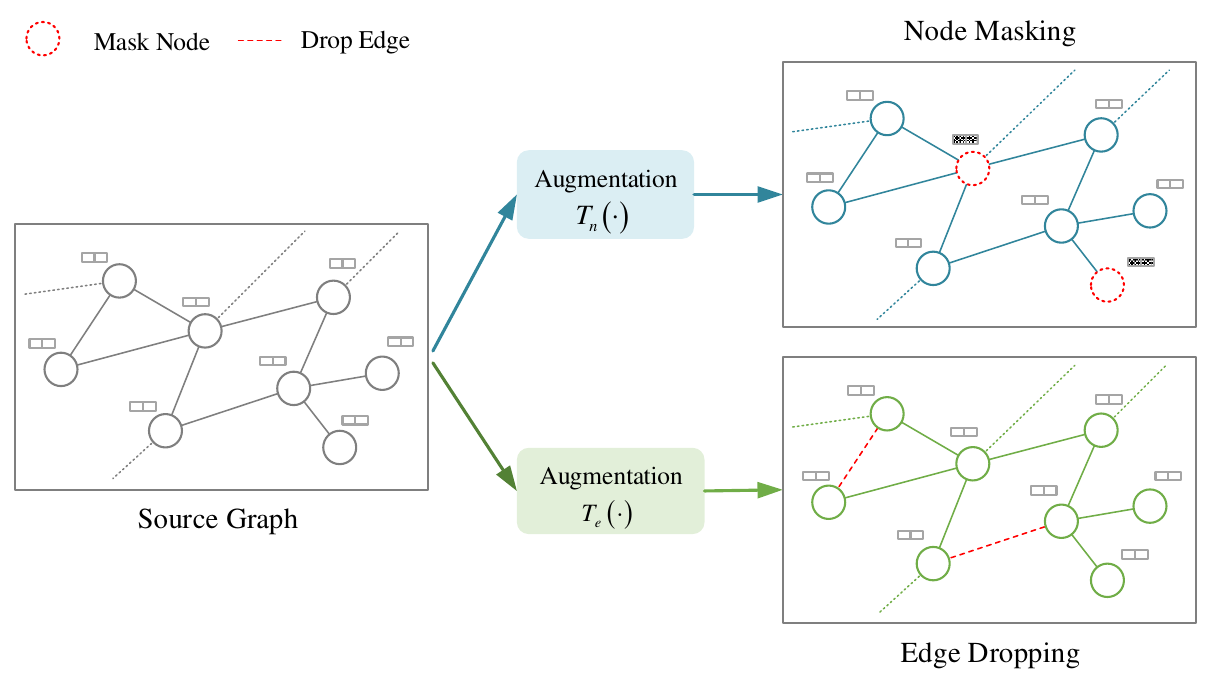}
	\caption{The examples of our major graph augmentation methods, including Edge Dropping,and Node Masking.}
	\label{Fig.3}
\end{figure}

Further, as shown in Fig. \ref{Fig.3}, to augment our samples, we apply a \textbf{primary node masking strategy} with a mask matrix $\left[M_0\right]$ for the online encoder. Node masking refers to the random masking features of some nodes on a given graph, either completely masking the feature vector or masking several parts of the feature vector, which can be formally defined as:
\begin{equation}
	\begin{aligned}
		t_i\left( X \right) =M_0\otimes X
	\end{aligned}
	\label{equ10}
\end{equation}
where $\left[M_0\right]$ is the mask matrix.

Then the online decoder through $\mathcal{H}_ o\left( \cdot \right)$ maps the embedding feature $H$ back to the feature vector $X^\prime$. As graphs, the amount of information reconstructed by the decoder is comparatively more insufficient than the multi-dimensional node features. Regular decoder or simple multi-layer perceptron (MLP) decoding expression ability is low, so we choose more expressive GNN as our decoder.

Additionally, for the target decoder, to better learn neighborhood information, we employ a \textbf{secondary edge dropping strategy}. Edge dropping is one of the most commonly used topology enhancement methods, which focuses on perturbing the graph adjacency matrix by randomly dropping some edges:
\begin{equation}
	\begin{aligned}
		t_i\left( A \right) =M_1\otimes A
	\end{aligned}
	\label{equ11}
 \end{equation}
where $M_1$ is the drop matrix.

After that, we take the generated latent representation from the online encoder and target encoder to contrastive learning, where the online decoder is used to conduct the reconstruction tasks. Through multiple rounds of training and optimization of joint losses, contrastive learning can learn potential discriminability by relationships between different images, and masked image modeling can  obtain high-level local features by reconstructing masked blocks. After possessing robust feature representation learning capability, only the online encoder is utilized for downstream age estimation.
\subsection{Training Objective}

\textbf{Reconstruction loss}: Corroborating the findings of \cite{hou2022graphmae}, MSE (mean square error) loss may not be enough for feature reconstruction while existing graph autoencoders (GAEs) have adopted MSE as their reconstruction criteria. Based on it, we employ the scaled cosine error (SCE) for our autoencoders computing reconstruction loss, which is formulated as follows:
\begin{equation}
    \begin{aligned}
        \mathcal{L}_{r c}=\frac{1}{\lvert \tilde{V}\lvert} \sum\left(1-\frac{x_i^T x_i^{\prime}}{\left\|x_i\right\| \cdot\left\|x_i^{\prime}\right\|}\right)^\gamma, \gamma \geqslant 1
    \end{aligned}
	\label{equ12}
\end{equation}
where $\tilde{V}$ denotes the masked subset nodes of $V$, $x_i$ is the primitive image features, $x_i^{\prime}$ is the reconstructed output. The scaling factor $\gamma$ is a hyper-parameter adjustable among different datasets and the loss is averaged over all masked nodes.

\begin{figure}
	\centering
	\includegraphics[width=0.48\textwidth]{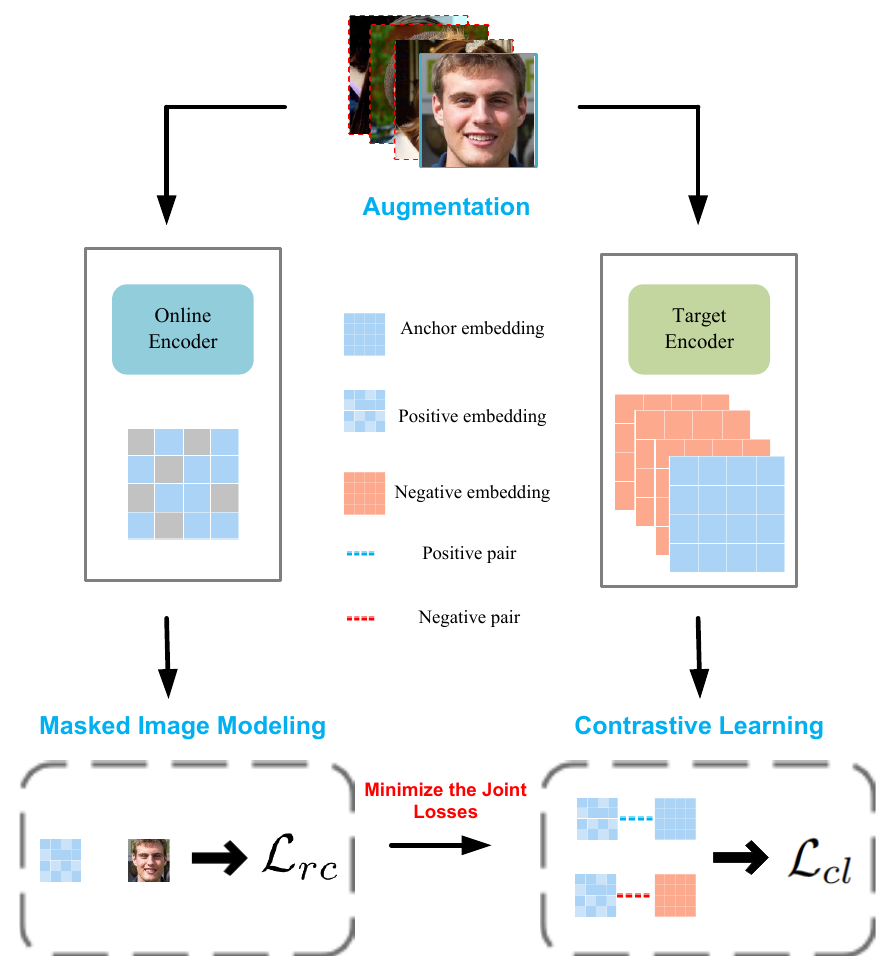}
	\caption{The illustration of training objective. MMGL adopts an asymmetric siamese network, utilizes graph augmentation methods and minimizes the joint losses to promote organic combination of the two mechanisms. }
	\label{Fig.4}
\end{figure}

\textbf{Contrastive loss}: In order to simulate the performance of real-life images under the influence of lighting, angle and other factors, we perform data augmentation by changing the grayscale and shifting the original image. Through data augmentation while preserving the content to form a positive sample, we generate two views ${I_i}$, $I_i^{+} \in \mathbb{R}^{H \times W \times C}$ of each image and respectively provide them to the online encoder and target encoder. And we store the other images in a batch as the negative samples for contrastive objective. Moreover, after encoding, we used the MLP projection head $\mathcal{P} \left( \cdot \right) : \mathbb{R} ^{N\times M}\mapsto \mathbb{R} ^{N\times M}$ to project the graph representation into the space where the contrastive loss is computed.

Our contrastive objective is based on the augmentation of the original image, following RELIC \cite{chuang2022robust}, introducing an invariance objective, which ensures that a data point and its positives and negatives are invariant in the expansion of that data point. To optimize the invariance, we define the Kullback-Leibler divergence between the probabilities of two augmented data views as the \textbf{invariance loss}, which is formulated as:
\begin{equation}
	\begin{aligned}
		KL\left( P\left( \mathcal{G} _i \right) \left\| P\left( \mathcal{G} _{i}^{+} \right) \right. \right) =sg\left[ \mathbb{E} _{P\left( \mathcal{G} _i,\mathcal{G} _{i}^{+} \right)}\log P\left( \mathcal{G} _i,\mathcal{G} _{i}^{+} \right) \right] \\ -\mathbb{E} _{P_{\left( \mathcal{G} _i,\mathcal{G} _{i}^{+} \right)}}\log P\left( \mathcal{G} _i,\mathcal{G} _{i}^{+} \right)
	\end{aligned}
	\label{equ2}
 \end{equation}
where  $P$ is the data probability distribution, $KL\left( \cdot \right)$ is the Kullback-Leibler (KL) divergence, $\mathcal{G}_{i}^{+}$ is the augmented sample of $\mathcal{G}_{i}$ , and $sg \left( \cdot \right)$ denotes the stop gradient, which can avoid degenerate solutions during optimization.

In addition to this, we managed to combine the usual contrastive loss InfoNCE \cite{he2020momentum} and modify it to fit the graph level shown as:
\begin{equation}
	\begin{aligned}
		\mathcal{L} _{cl}=-\alpha \log \frac{\exp \left( \mathcal{G} _i\cdot \mathcal{G} _{i}^{+}/\tau \right)}{\sum{_{j=0}^{K}\exp \left( \mathcal{G} _i\cdot \mathcal{G} _j/\tau \right)}} \\ +\eta KL\left( P\left( \mathcal{G} _i \right) \left\| P \right. \left( \mathcal{G} _{i}^{+} \right) \right)
	\end{aligned}
	\label{equ13}
 \end{equation}
where $K$ is the batch size, $\tau$ is the temperature constant, $\mathcal{G}_{i}^{+}$ is the positive sample obtained by data augmentation of $\mathcal{G}_{i}$ and $\mathcal{G}_{j}$ is the negative sample of $\mathcal{G}_{i}$ originating from different images in the queue. The relative importance of the contrastive loss and invariance loss is weighted by the $\eta$ scalar.

Overall, Fig. \ref{Fig.4} explicitly shows the training architecture. The training objective is a weighted loss function by combining reconstruction loss $\mathcal{L} _{rc}$ and contrastive loss $\mathcal{L} _{cl}$ defined as:
\begin{equation}
	\begin{aligned}
		\mathcal{L} _{MC}=\mu \mathcal{L} _{rc}+\left( 1-\mu \right) \mathcal{L} _{cl}
	\end{aligned}
	\label{equ15}
  \end{equation}

The entire inference process of the MMCL pseudocode is contained in Algorithm 1.

\begin{table}[htbp]
	\centering
	\renewcommand\arraystretch{1.5}
	\setlength{\tabcolsep}{1pt}{
		\resizebox{\linewidth}{!}{
		\begin{tabular}{l}
			\toprule
			\textbf{Algorithm 1} Multi-view Mask Contrastive Learning (MMCL).
			\\ \hline
			\textbf{Input:} images $\mathcal{I} =\left\{ I_1, I_2, ... , I_n \right\}$, mask matrix $M_0$,drop \\ matrix $M_1$ and  hyper-parameters $m$, $\gamma$,  $\eta$, $\mu$;
			\\
			\textbf{Output:} The Online Encoder with parameters $\varTheta_o$.  \\
			1: \textbf{for} epoch=$1, \ldots, K$ \textbf{do}\\
			2:  \qquad  \textbf{for} sampled minibatch ${{\cal G}_i}$, $i$=$1, \ldots, N$  \textbf{do} \\
			3: \qquad \qquad data augmentation, $t\left( \mathcal{G} _{i}^{+} \right) \sim \tau \left( \mathcal{G} _i \right) $. \\
			\qquad \qquad \quad  \textbf{// primary node masking.} \\
			4: \qquad \qquad $\mathcal{G} _i\sim t_i\left( X \right) =M_0\otimes X$. \\
			\qquad \qquad \quad  \textbf{// secondary edge dropping.} \\
			5: \qquad \qquad  $\mathcal{G} _{i}^{+}\sim t_i\left( A \right) =M_1\otimes A$. \\
			\qquad \qquad \quad  \textbf{// graph encoder.} \\
			6: \qquad \qquad online encoder, $H_i=\mathcal{F} _o\left( A,t_i\left( X \right) \right)$. \\
			7: \qquad \qquad target encoder, $H_{i}^{+}=\mathcal{F} _t\left( t_i\left( A \right) ,X \right)$. \\
			\qquad \qquad \quad  \textbf{// graph decoder.} \\
			8: \qquad \qquad target decoder, ${{\cal G}_i}^\prime  = {{\cal H}_o}\left( {A,{H_i}} \right)$. \\
			\qquad \qquad \quad  \textbf{// parameter update.} \\
			9: \qquad \qquad calculate loss ${L_{rc}}$ by using Eq. (12). \\
			10: \qquad \quad \: calculate loss ${{\cal L}_{cl}}$ by using Eq. (14). \\
			11: \qquad \quad \: calculate loss ${{\cal L}_{MC}}$ by using Eq. (15). \\
			12: \qquad \quad \: update parameters $\varTheta_o$ by using Eq. (9). \\
			13: \qquad \textbf{end for} \\
			14: \textbf{end for} \\
			15: Return the online encoder network. \\
			\bottomrule
		\end{tabular}}
	}
\end{table}

\subsection{Age Estimation}
In the age estimation stage, we do supervised training to evaluate the representations from the online encoder. Our age estimation framework consists of three components: feature extraction, age grouping, and final prediction.

\textbf{Feature Extraction}: For the feature extraction model, we first pre-train the model in an unsupervised manner. We then use the corresponding attribute labels to train and fine-tune the feature extraction model. For age estimation, we then combine the features extracted by the fine-tuning network into a feature matrix to train the ELM model.

The classical extreme learning machine (ELM) is a single implicit layer feedforward neural network with high efficiency, high correctness, and strong generalization performance, which the output equation of the $i$-th hidden layer node is:
\begin{equation}
	\begin{aligned}
		h_i\left( x \right) =G\left( a_i,b_i,x \right)
	\end{aligned}
	\label{equ16}
 \end{equation} where $a_i$ denotes the $i$-th hidden node weight vectors, $b_i$ denotes the $i$-th hidden node weight biases, and $G\left( a_i,b_i,x \right)$ denotes a nonlinear piecewise continuous function.

The output is weighted and summed according to whether the task is a classification or a regression, single or multi-classification, which is formulated as:
\begin{equation}
	\begin{aligned}
		f\left( x \right) =\sum_{i=1}^L{\beta _iG\left( a_i,b_i,x \right) =\beta H}
	\end{aligned}
	\label{equ17}
 \end{equation}
where $\beta$ denotes the output weights’ vector, and $H$ denotes the output matrix of the hidden layer.

That learning goal to minimize the loss function can be written as follows:
\begin{equation}
	\begin{aligned}
		\min : \lambda \left\| f\left( x \right) -T \right\| _{\mu}^{\sigma _1}+\left\| \beta \right\| _{\nu}^{\sigma _2}
	\end{aligned}
	\label{equ18}
\end{equation} where $\sigma_1>0$, $\sigma_2>0$, and $\mu$, $\nu$ = 0,  $\frac{1}{2}$, 1, ···, $+\infty$. $\lambda$ is the parameter that weighs between these two, $T$ is the training data target matrix.

\begin{figure}
	\centering
	\includegraphics[width=0.48\textwidth]{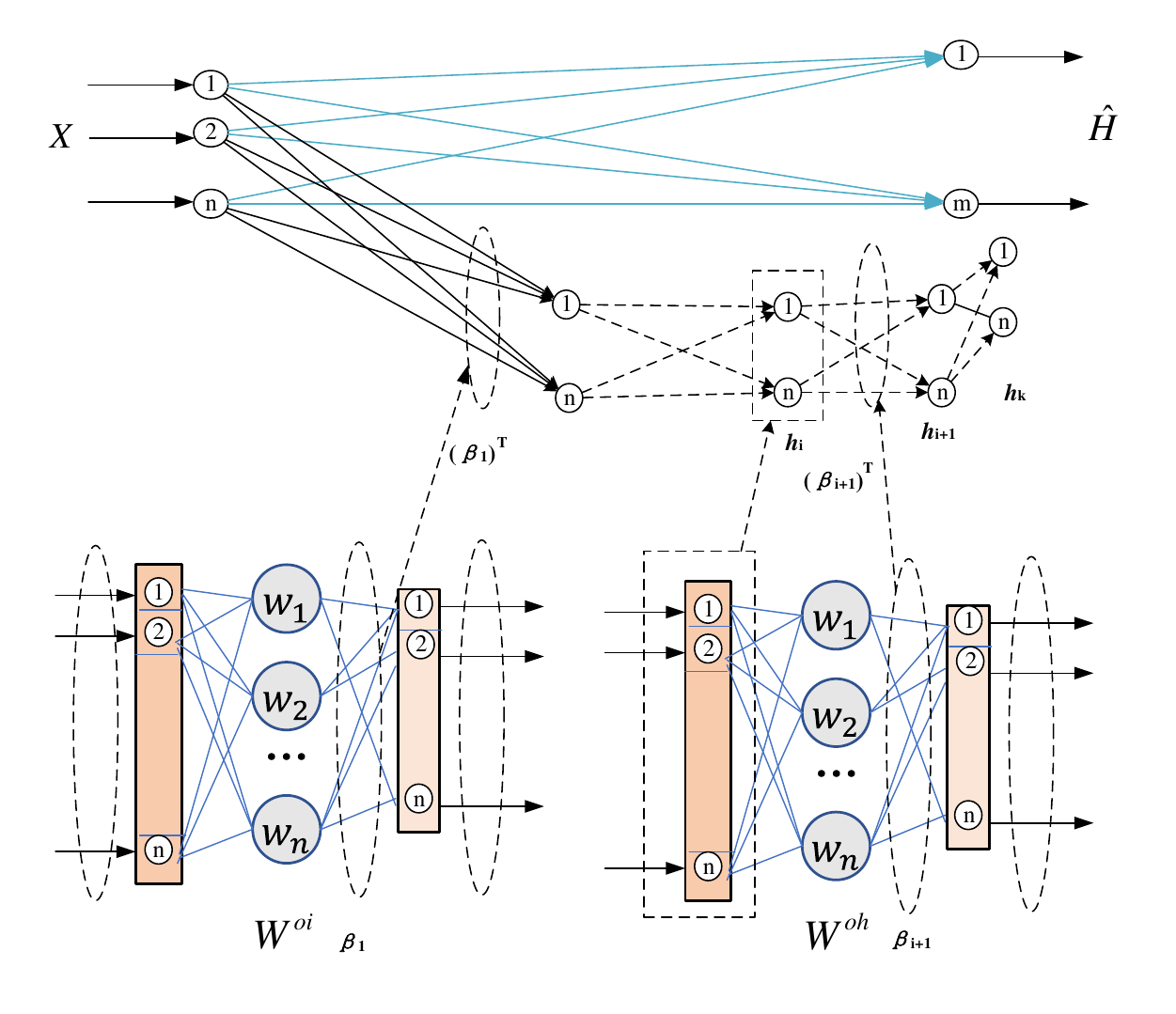}
	\caption{The illustration of multilayer ELM based on identity mapping.}
	\label{Fig.5}
\end{figure}

However, there are some shortcomings in classical ELMs, which have limited learning ability for high-dimensional complex data features and are prone to overfitting based on empirical risk minimization. Multilayer ELM stacks multiple ELM into a multilayer ELM autoencoder (ML-ELM), which is suitable for data dimensionality reduction, image noise reduction, and representation of high-dimensional complex abstract features. The dimensionality reduction of features helps improve classification accuracy and reduce the error rate. Further, this paper adopts ML-IELM (multilayer ELM based on identity mapping) so that the output neurons can not only obtain the re-coding of data through the nodes of the last hidden layer but also obtain the data information directly from the input neurons through the constant mapping, which increases the richness and comprehensiveness of the data obtained by the output neurons of the network, and thus achieves the purpose of improving the generalization and accuracy of the network. The structure of multi-layer extreme learning machine with identity mapping is shown in Fig. \ref{Fig.5}.

The mathematical model of ML-IELM is:
\begin{equation}
	\begin{aligned}
		{\cal F}\left( x \right) = {\cal F}\left\{ {{W^{oi}}X + \left( {{W^{oh}}\hat H} \right) + C} \right\}
	\end{aligned}
	\label{equ17}
 \end{equation}
where ${{W^{oi}}}$ denotes the weight matrix between the output neuron and the input neuron, ${{W^{oh}}}$ denotes the matrix of weights between the last implicit layer and the output layer, ${\cal F}\left( . \right)$ denotes the identity mapping function, $X$ indicates the input data, ${\hat H}$ denotes the output matrix of the last hidden layer and $C$ denotes the bias term.

The goal of the ML-IELM network is to compute the optimal weight matrix from the hidden node to the output node so that the network output at that weight value minimizes the error between the output and the desired output value. The optimization model of the network is defined as:
\begin{equation}
	\begin{aligned}
		\begin{array}{l}
\min :\lambda \left\| {{\cal F}\left( x \right) - T} \right\|_\mu ^{{\sigma _1}} + \left\| \beta  \right\|_\nu ^{{\sigma _2}}\\
 = \lambda \left\| {{\cal F}\left( {{W^{oi}}{W^{oh}}C} \right)\left[ {\begin{array}{*{20}{c}}
X\\
{\hat H}\\
I
\end{array}} \right] - T} \right\|_\mu ^{{\sigma _1}} + \left\| \beta  \right\|_\nu ^{{\sigma _2}}
\end{array}
	\end{aligned}
	\label{equ17}
 \end{equation} where $\sigma_1>0$, $\sigma_2>0$, and $\mu$, $\nu$ = 0,  $\frac{1}{2}$, 1, ···, $+\infty$. $\lambda$ is the parameter that weighs between these two, $T$ is the training data target matrix and $I$ is the unit matrix.

\textbf{Age Grouping}: First, we use ML-IELM to train the classifier, classify faces into different age groups, and reduce the task difficulty of subsequent regressors. During training, we obtain the average weights and biases of the hidden layers of the ELM classifier model.

\textbf{Final Prediction}: After using ML-IELM classifiers to divide the age groups into smaller groups, we utilize the smaller age groups with accurate age labels to train the ML-IELM regressor. During training, we obtain the average weights and biases of the hidden layer of the ELM regressor. Then we make a final age prediction via the trained ML-IELM regressor.

\section{Experiments}
In this section, we present experimental details and demonstrate the effectiveness of the proposed representation learning model MMCL-GCN. The feature extraction module MMCL (Multi-View Mask Contrastive Learning) in the MMCL-GCN model can be trained unsupervised. Therefore, we first pre-train MMCL on ImageNet-1K \cite{deng2009imagenet} dataset in a self-supervised manner, where all frames are extracted. We then further train the MMCL-GCN model on the IMDB-WIKI \cite{rothe2015dex} dataset in a supervised manner to evaluate the learned representation for age estimation. After that, we conduct experiments in the popular age estimation datasets and compare them with the baselines. Taking into account the characteristics of different datasets, we employ varied criteria to better accurately assess the model capabilities. The proposed architecture is trained on an NVIDIA Tesla V100 and implemented with the publicly available Caffe codes. We resize the original images to 256× 256 pixels, and then select a 224 × 224 crop from the center or the four corners from the entire processed image. Moreover, we leverage different dropout measures to reduce the risk of overfitting. In the end, we perform ablation studies on each component to investigate the individual functions in our framework.

\subsection{Dataset}
Experiments are carried out on a number of age estimation benchmark datasets: IMDB-WIKI, Morph II \cite{ricanek2006morph}, Adience Benchmark \cite{eidinger2014age}, LAP-2016 Dataset \cite{escalera2016chalearn}. Our feature extraction module MMCL is pretrained on ImageNet-1K and IMDB-WIKI following previous work, and evaluated on the Adience Benchmark, Morph II and LAP-2016 datasets.

\textbf{IMDB-WIKI}: This dataset is one of the largest age datasets available. The sources include 20,000 individuals from IMDB (a celebrity website) and Wikipedia with 460723 and 62328 images, respectively, because it contains more noise, e.g., multiple faces in one image. Therefore, we only use it for pre-training and fine-tuning.

\textbf{MORPH-II}: The dataset is the most popular age estimation dataset available, containing 55,134 images of 13,000 people's faces, with photos collected by people aged 16-77 years, with an average age of 33 years. In addition to age, other attributes about the person was recorded, such as race and gender. Referring to a widely used evaluation protocol \cite{duan2017ensemble}, the dataset was randomly divided into two non-overlapping parts, the training dataset (80$\%$) and the test dataset (20$\%$).

\textbf{Adience Benchmark}: The dataset consists of 26,580 images of 2,284 people, divided into eight intervals using interval annotation. The images were taken in real scenes and are heavily influenced by noise, pose and lighting, implying a solution to the problem of age and gender detection in the real life. Both the raw data and corrected faces are available on the website.

\textbf{LAP-2016 Dataset}: The LAP (Look At People) competition was held in 2016 and consisted of 7,591 images, with each sample labeled with mean $\mu$ and standard deviation $\sigma$. The LAP-2016 dataset was relatively distributed across ages 20-40, with fewer datasets in the 0-15 and 65-100 intervals, so we pre-trained the network on the ImageNet-1K and IMDB-WIKI dataset and then just fine-tuned the aging network. The dataset is divided into 4,113 training images, which 1,500 for validation and 1978 for testing.

\subsection{Unsupervised Training}
Following the standard architecture, we use ViG-S \cite{han2022vision} as the encoders and a 2-layer shallow GCN as the decoder. We utilize ImageNet-1K for pre-training with unsupervised manner and each sample generate a view by data augmentation such as greyscaling and shifting. Although the cotrastive objective requires a sufficient number of negative examples, we do not need large batches by default using queue and therefore set the batch size to 1,024. For large datasets, we use cumulative gradients to simulate large batch sizes. By default, we pre-trained the model for 1,600 epochs with 50 warm-up epochs. The learning rate is 1.5e$-$4 with 0.01 weight decay. The mask ratio is set to 0.50 and drop ratio is set to 0.20 by default.

We compare the representation learning performance of MMCL with the computer vision baselines. For contrastive SSL, MoCo \cite{he2020momentum} utilized a queue instead of a memory bank to store negative samples and replaces the constraint term in the loss with a momentum encoder for better results. SimCLR \cite{chen2020simple} used more data augmentation, larger batchsize and a nonlinear projection header to maximize the agreement expressed in the plan. For generative SSL, MAE \cite{he2022masked} employed random mask to process the input image blocks and reconstructed the masked image patches for training. CAE \cite{chen2024context} introduced latent context regressor to predict the hidden representation of masked patches based on the hidden representation of visible patches and then decoded and reconstructed it.

\begin{figure}
	\centering
	\includegraphics[width=0.48\textwidth]{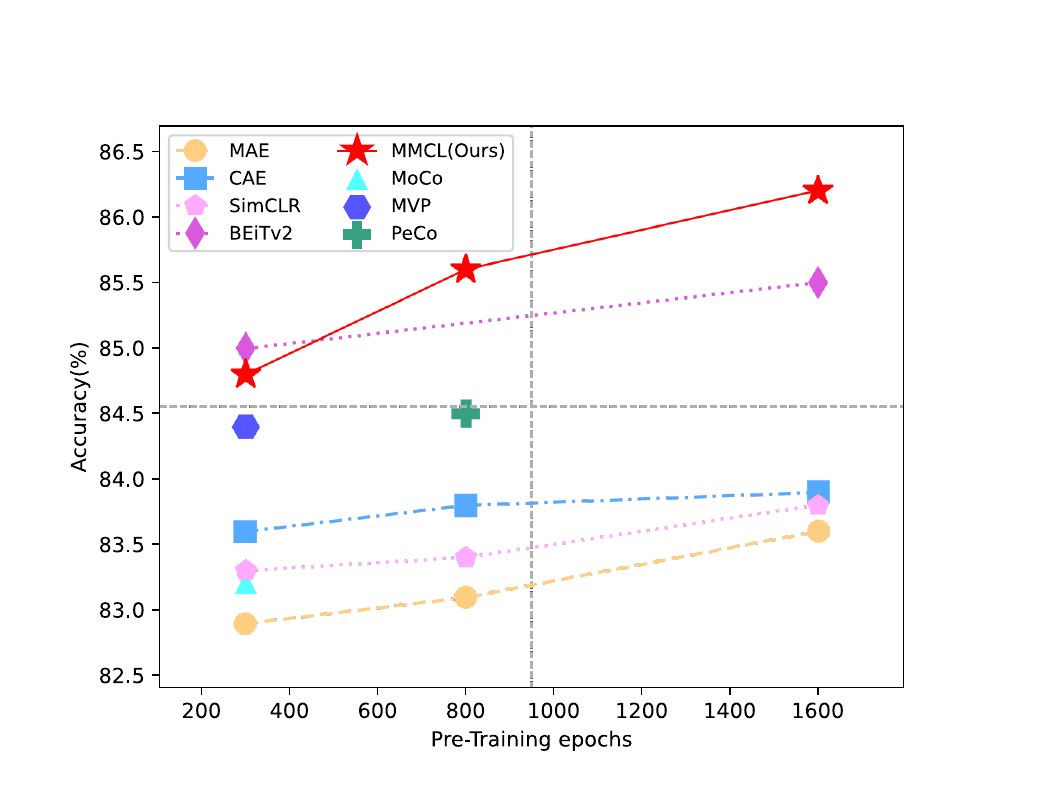}
	\caption{The pre-training performance of MMCL with state-of-the-art methods in terms of top-1 accuracy on ImageNet-1K.}
	\label{Fig.6}
\end{figure}

Then we evaluate our graph representation learning pre-trained model using top-1 accuracy on ImageNet-1K classification. As shown in Fig. \ref{Fig.6}, our MMCL significantly improves the representation learning between masked image modeling and contrastive learning, which outperforms the state-of-the-art. We achieve 86.2$\%$ with ViG-S not only outperform the contrastive methods such as MoCo, but also competitive with the SOTA methods such as MAE (83.6$\%$) and CAE (83.9$\%$).

\subsection{Supervised Training}
\textbf{Pre-training}: In supervised training, we first load the pre-trained weight and train the entire model on IMDB-WIKI with both images and the corresponding labels. By default, the learning rate is 1.5e-3 with 0.01 weight decay.

\textbf{Fine-tuning}: For the age estimation tasks, we train the facial datasets to obtain strong and robust feature extraction model. The networks are initialized with the weights gained from training on the ImageNet and IMDB-WIKI. Following the experiment settings in \cite{duan2018hybrid}, we evaluate the proposed method on Morph-II, LAP-2016 dateset and an unconstrained dataset Adience Benchmark. For the networks, we set the base$\_$lr as 0.01, momentum as 0.9, and gamma as 0.1. The weight$\_$decay is set as 0.0005.  For ELM classification and regression, $\beta$ is set as $\{$ $10^{-5}$, $10^{-4}$, ... , $10^2$, $10^3$$\}$ and $L$ is set as $\{$1800,2000, ... , 4000$\}$.

\textbf{Optimization}: We use AdamW as the default optimizer for pre-training, fine-tuning and age estimation, regarding age estimation, AdamW is still preferred, but we also use stochastic gradient descent (SGD) and LARS. We use the linear scaling rule: $lr = base\_lr \times batch\_size/256$ to set the learning rate. We also adopted the cosine learning rate schedule with a warmup of 50 epochs.

\subsection{Evaluation Criteria}

\textbf{MAE(Mean Absolute Error)}: The mean absolute value of the deviation of each sample's label from its corresponding estimate is a useful measure of the accuracy of the ELM regressor. In general, the mean absolute error (MAE) for age prediction can be calculated as follows:
  \begin{equation}
	\begin{aligned}
        MAE=\frac{1}{N} \sum_{i=1}^N\lvert{x_i-y_i}\rvert
	\end{aligned}
	\label{equ19}
 \end{equation} where $x_i$ denotes the true label, $y_i$ is the predicted value, and $N$ represents the number of testing samples.

 \textbf{CS(Cumulative Score)}:  The cumulative score is more appropriate when training samples are available at essentially every age. The larger the cumulative score, the better the age estimation performance, which is defined as:
\begin{equation}
	\begin{aligned}
		CS\left( L \right) =\left(n_{e}\leqslant L/N \right) \times 100\%
	\end{aligned}
	\label{equ20}
 \end{equation} where $n_{e}\leqslant L$ expresses the number of test images whose absolute error $\epsilon$ of the age estimation is not larger than $L$ years.

\textbf{Normal Score($\epsilon$-error)}: Calculate the proportion of samples that are inaccurately predicted as a percentage of all samples. Owing to the LAP-2016 dataset was labelled through several annotators, the performance of age estimation can be more accurately measured by taking into account the following factors. The smaller the $\epsilon$-error, the better the performance of the age estimation classifier. By fitting a normal distribution with mean ${\mu}$ and standard deviation ${\sigma}$ of the annotations for each sample, the $\epsilon$-score is calculated as follows:
\begin{equation}
	\begin{aligned}
		\epsilon =1-e^{-\frac{\left( x-\mu \right) ^2}{2\sigma ^2}}
	\end{aligned}
	\label{equ21}
 \end{equation}
Therefore, $\epsilon\in \left(0, 1\right)$, and 1 is worst case while 0 denotes best case.

\subsection{The State-of-the-Art Baselines Comparison}

We compare our proposed architecture with various state-of-the-art estimation baselines. We classify competing methods into bulky models and compact models based on their model sizes. For bulky models, DEX \cite{rothe2015dex} was pre-trained in ImageNet based on VGG-16, defining the regression problem as a classification problem with softmax expectation refine, which achieved improvement over direct regression training. BridgeNet \cite{li2019bridgenet} divided the data space by a local regressor to handle heterogeneous data, and utilized a gated network to learn the continuous weights used by the local regressor. For compact models, SSR-Net \cite{yang2018ssr} adopted a coarse-to-fine strategy to adjust the predicted age stage by stage dynamic range. C3AE \cite{zhang2019c3ae} proposed a multi-scale feature stitching approach to compensate for the small model fitting problem by two-level serial supervision.

\textbf{Age Estimation on the MORPH-II}: Our method combines the MMCL with ML-IELM and compares the performance of MORPH-II with the state-of-the-art baseline methods. The robustness and effectiveness of MMCL-GCN are analyzed in terms of the MAEs. Table \ref{results1} presents the detailed results, which list the backbone networks and MAE.

Our method achieves the best performance among previous works. For bulky models, DEX achieved good results attribute to the recalibration of face image rotation and pre-training on IMDB-WIKI. BrigeNet jointly trained local regressors and gated networks that could be easily integrated into end-to-end models with any other deep neural networks. RANG and EGroupNet took into account race, gender factors affecting age prediction. Good results have also been obtained recently for some lightweight compact models. SSR-NET proposed dynamic soft stagewise that can be panned and scaled, and C3AE is efficient based on cascading contexts. However, the former is scalable but has more redundant parameters, while the latter is compact but difficult to handle age estimation in complex environments. Our model uses a medium-scale graph neural network and two powerful ML-IELM with multi-factor feature extraction, which is both robust, compact and scalable.

\begin{table}
	\caption{Comparison with the state-of-the-art methods on Morph II.}
	\renewcommand\arraystretch{1.4}
	\setlength\tabcolsep{2mm}
	\centering
		\begin{tabular}{l | c c c}
			\hline
			Method & Backbone & MAE & Model size \\
			\hline
			Ranking\cite{chang2010ranking}	 & - & 2.96 & 2.2GB \\
			DEX\cite{rothe2015dex} & VGG-16 & 2.68	& 530MB \\
			MV\cite{pan2018mean} & VGG-16 & 2.41 & 530MB \\
			BrigeNet\cite{li2019bridgenet} & VGG-16 & 2.35 & 530MB \\
			RANG\cite{duan2018hybrid} & CNN & 2.61 & -\\
            EGroupNet\cite{duan2020egroupnet} &	- & 2.48 & - \\
			DLDLF\cite{shen2019deep} & VGG-16 & 2.19 & 530MB\\
			\hline
            C3AE\cite{zhang2019c3ae} & - & 2.78	& 0.25MB\\
            SSR-NET\cite{yang2018ssr} & - & 3.16 & 0.32MB\\
            \hline
            \textbf{MMCL-GCN}	& \textbf{ViG-S} & \textbf{2.13} & \textbf{47.8MB}\\
			\hline
	\end{tabular}
	\label{results1}
\end{table}

We were able to achieve such results attributed to: i) the strong feature extraction capability in a smaller volume based on graph neural networks, combined with contrastive learning and masked image modeling. ii) the effective graph augmentation strategy and the excellent loss function designed that enable the representation learning beyond convolutional neural networks. iii) the integrated  multiple attributes of race, gender and age, and the use of classification before regression, which greatly improve the prediction accuracy.

 \textbf{Age Estimation on the LAP-2016 Dataset}: We further provide the results of the MMCL-GCN evaluated on LAP-2016 dataset. The test error is employed for evaluation owing to this dataset was labelled through several
 annotators, so the performance of age estimation can be more accurately measured by this criterion. Table \ref{results2} presents the final results. Our method achieves 0.2157 $\epsilon$-score outperform the state-of-the-art methods on LAP-2016 dataset. This is probably attributed to the powerful feature extraction capability of our graph representation learning model and the more accurate predictions resulting from the ML-IELM classification and regression strategy.

\begin{table}
	\caption{Comparison with the state-of-the-art methods on LAP-2016. The data in the table are mainly from \cite{agbo2021deep}.}
	\renewcommand\arraystretch{1.4}
	\setlength\tabcolsep{2.5mm}
	\centering
		\begin{tabular}{l | c | c}
			\hline
			Method & MAE & $\epsilon-$error \\
			\hline
			4-fusion framework & - & 0.3738 \\
			RANG\cite{duan2018hybrid} & - & 0.3679 \\
			EGroupNet\cite{duan2020egroupnet} & - & 0.3578\\
			VGG+Revised Contrative Loss & - & 0.3679\\
			AGEn & 3.82 & 0.3100\\
            GNN+Mean-Varian Loss & - & 0.2867 \\
			AL  & - & 0.2859\\
            DLDL-V2 & 3.452 & 0.267\\
            Children-specialized CNN & - & 0.2411\\
            Soft-Ranking & - & 0.232\\
           \textbf{MMCL-GCN} & -	 & \textbf{0.2157}\\
			\hline
	\end{tabular}
	\label{results2}
\end{table}

 \textbf{Age Estimation on the Adience Benchmark}: We use an unconstrained dataset to vetisfy the performance of our model, and the average accuracy is more suitable to evaluate it for which is more difficult to distinguish races because the images were not manual filtered initially. Table \ref{results3} shows the age estimation results on the Adience Benchmark. We obtained an average accuracy of 0.7031, which shows that our model is competitive even in complex real-world environments. This may be attributed to our strong representation learning model and the benefit of pre-training on large datasets.

\begin{figure*}
	\centering
	\includegraphics[width=1\textwidth]{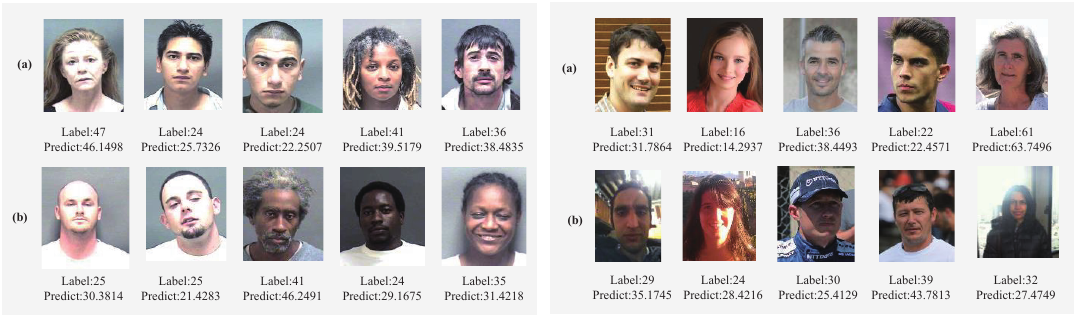}
	\caption{The examples of satisfactory estimation (a) and unsatisfactory estimation (b) on Morph-II (left) and Adience benchmark(right). Despite the mistakes of our model in some pictures, the deviations of unsatisfactory predictions from the actual values are basically in the same age group compared to other methods \cite{duan2018hybrid, yang2018ssr}, and we are also better in predicting faces in the face of some complex environments.}
	\label{Fig.7}
\end{figure*}

Some samples of age prediction results are provided in Fig. \ref{Fig.7}, the images on the left are from the Morph-II and the right are from the LAP-2016 dataset. (a) shows the satisfactory predictions, illustrating the effectiveness of GNNs for feature extraction and the reliability of our method. (b) Shows some unsatisfactory predictions, possibly due to face misdetection, alignment errors and complex environment.

\begin{table}
	\caption{Comparison with the state-of-the-art methods on Adience Benchmark.}
	\renewcommand\arraystretch{1.4}
	\setlength\tabcolsep{7mm}
	\centering
		\begin{tabular}{l | c }
			\hline
			Method & Average Accuracy \\
			\hline
			DEX\cite{rothe2015dex}  & 0.64$\pm$0.042 \\
			Best from\cite{duan2020egroupnet} & 0.5071$\pm$0.051  \\
			CNN+ELM\cite{gurpinar2016kernel} & 0.5127$\pm$0.0497 \\
			RANG\cite{duan2018hybrid} & 0.6649$\pm$0.0508\\
            EGroupNet\cite{duan2020egroupnet} &	0.6978$\pm$0.0713 \\
            \textbf{MMCL-GCN} & \textbf{0.7031$\pm$0.0419}\\
			\hline
	\end{tabular}
	\label{results3}
\end{table}

\begin{table}
	\caption{The ablation study of graph augmentation strategies.}
	\renewcommand\arraystretch{1.4}
	\setlength\tabcolsep{1.5mm}
	\centering
		\begin{tabular}{l | c c c }
			\hline
			index & Node mask & Edge drop & Accuracy\\
			\hline
			1 & $\times$  & $\times$ & 84.5 \\
			2 & $\times$  & \checkmark & 84.9\\
            3 & \checkmark  & $\times$ & 85.1\\
            \textbf{4(MMCL-GCN)} & \checkmark  & \checkmark & \textbf{85.6}\\
			\hline
	\end{tabular}
	\label{results4}
\end{table}

\subsection{Ablation Studies}
To analyze the effectiveness of the key components and methodological strategies in the architecture and to validate the design we adopted at MMCL-GCN, we perform a series of ablation experiments, and our performance is based on the model with 800 pre-training epochs.

\textbf{The Effects of graph augmentation strategies}: We study the influence of the graph augmentation strategy, and Table \ref{results4} shows the result of primary node masking and secondary edge dropping strategies. The results manifest the ability of using a graph augmentation strategy alone to achieve a 1$\%$ to 2$\%$ improvement, and a more significant improvement of 2$\%$ to 6$\%$ when the two are used together. Because our structure is asymmetric, the organic combination of contrastive learning and masked image modeling, using graph augmentation strategies can maximize the advantages of both, obtaining both local and global features. The primary node masking
strategy is more helpful to learn local potential information, and the the secondary edge dropping strategy is more helpful to learn structural features for global capture, thus making the representations better compatible and complementary.

\begin{figure}
	\centering
	\begin{subfigure}
		\centering
		\includegraphics[width=0.48\textwidth]{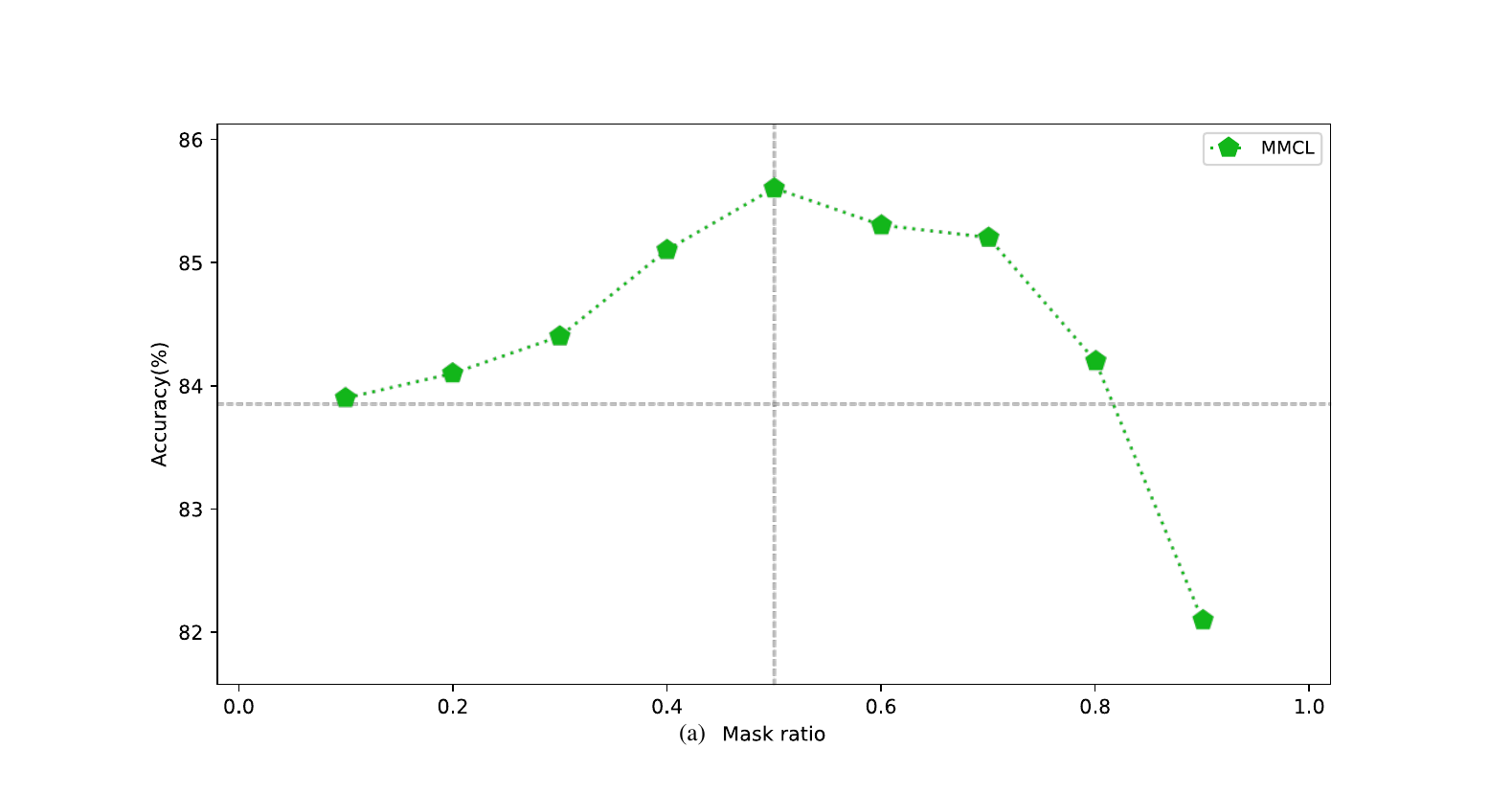}	
		\label{Fig.8(a)}
	\end{subfigure}
	\centering
	\begin{subfigure}
		\centering
		\includegraphics[width=0.47\textwidth]{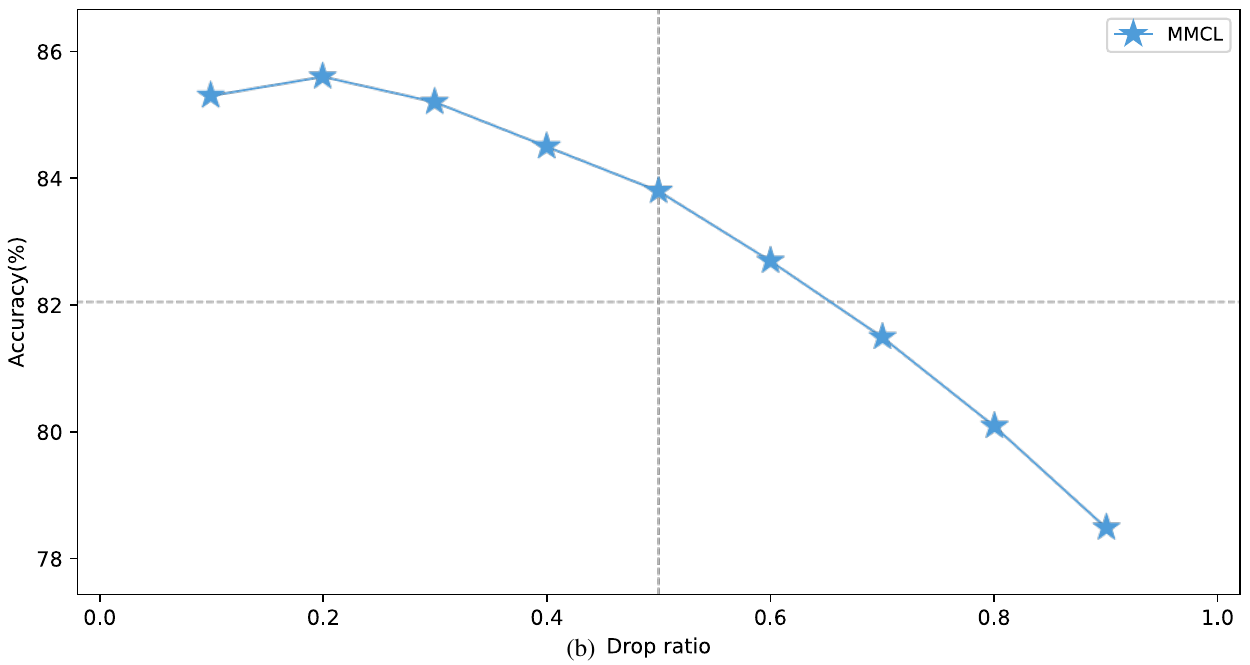}	
		\label{Fig.8(b)}
	\end{subfigure}
	\caption{The ablation study of mask ratio and drop ratio.}
	\label{Fig.8}
\end{figure}

\textbf{The Effects of mask ratio and drop ratio}: We further investigate the effect of node mask ratio and edge drop ratio on representation learning. For most cases, images have more redundant information and a higher mask ratio is more helpful for the model to learn features. Fig. \ref{Fig.8} (a) shows that in ImageNet-1K, the optimal mask ratio is 0.5${\sim }$0.75. However, in the facial datasets, too high a mask ratio degrades the performance. Further, due to the specificity of our structure, contrastive learning prefers more complete images for supervision, and too high mask ratio is detrimental to it. Therefore, our solution adopts a compromise, where the optimal mask ratio is 0.25${\sim }$0.5 in the facial datasets. We default the mask ratio to 0. 5 which the efficiency gain compensates for the small performance loss.

Fig. \ref{Fig.8} (b) shows the performance of different edge drop ratios. In most cases, our model maintains good capability, keeping a low edge drop ratio of 0.15${\sim }$0.2 leads to a large improvement in model performance, while there is a significant decrease in accuracy as the ratio increases. Probably, the feature information of the graph is more sensitive to edges, and too high drop ratio can seriously damage the graph structure. In our experiments, we set the default drop ratio to 0.2.

\textbf{The Effects of contrastive and reconstruction losses}: We compare our MMCL-GCN loss function with baselines MAE and MoCo. Table \ref{results5}  demonstrates that by jointly training the optimized contrastive loss and reconstruction loss, we end up with a smaller target loss. This motivates us that masked autoencoder and contrastive learning are not only compatible training targets, they are also complementary. By optimizing the loss function and joint training optimization, we obtain a gain of 1${\%}{\sim }$2.5${\%}$.

\begin{table}
	\caption{The ablation study of contrastive and reconstruction losses.}
	\renewcommand\arraystretch{1.4}
	\setlength\tabcolsep{3mm}
	\centering
	\begin{tabular}{l | c c }
		\hline
		Method & Contrastive loss & Reconstruction loss\\
		\hline
		MoCo & 8.714  & — \\
		MAE & —  & 0.1383\\
        \textbf{MMCL} & \textbf{8.529$\left(\downarrow\right)$} & \textbf{0.1349$\left(\downarrow\right)$} \\
		\hline
	\end{tabular}
	\label{results5}
\end{table}

\textbf{The Effects of ML-IELM modules}: Combining MMCL and ML-IELM is the key to performing robust age estimation. The powerful and effective MMCL is used for face feature extraction, and the trained ML-IELM is applied for age grouping before regression to obtain the final age prediction results. In a series of experiments based on the same parameter settings as the previous experiments, we change the choice of classifiers and regressors while following the same age prediction strategy.

\begin{figure}
	\centering
	\includegraphics[width=0.48\textwidth]{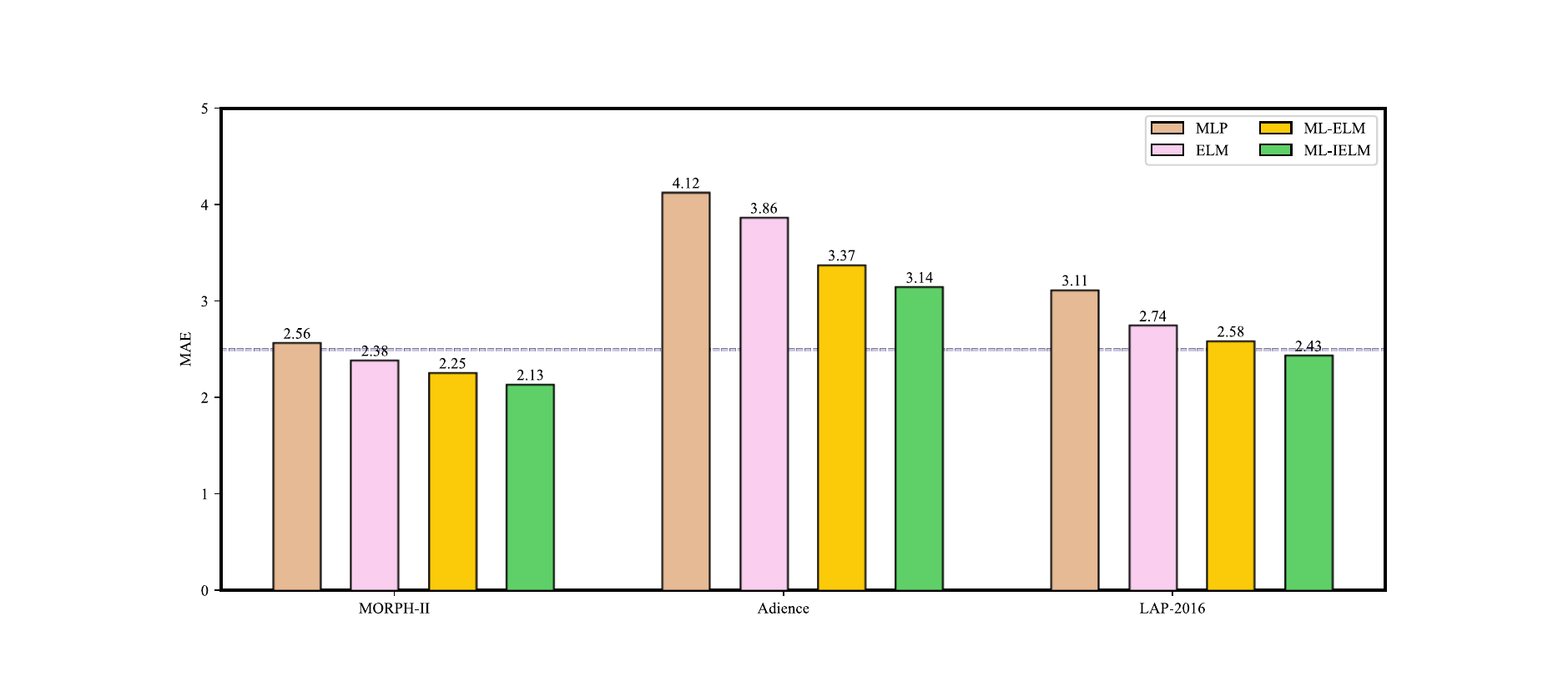}
	\caption{The ablation study of ML-IELM modules.}
	\label{Fig.9}
\end{figure}

We compare the simple three-layer MLP, the classical ELM, and the widely used ML-ELM by applying the proposed ML-IELM to three facial datasets using MAE (Mean Absolute Error) as the evaluation criterion. The comparison results in Fig. \ref{Fig.9} show that the ML-IELM proposed in this paper has a much lower age prediction error and better generalization performance. Compared with MLP, ELM network architecture is simpler and more efficient. Because ELM has limited learning ability for high-dimensional complex data features and is prone to overfitting based on empirical risk minimization, ML-IELM is further improved and optimized based on ML-ELM, which results in a better overall performance. Although the improved ML-IELM does not improve accuracy on the MORPH-II dataset, it has more obvious improvement on the unrestricted dataset of Aidence, i.e., it is reliable in the face of more complex and rich data features.

\section{Conclusion}
This paper has proposed a Multi-view Mask Contrastive Learning Graph Convolutional Neural Network for Age Estimation (MMCL-GCN). This method includes two stages: the feature extraction stage and the age estimation stage. At the graph level, MMCL-GCN leverages contrastive learning to fuse masked image modeling for obtaining a robust and efficient feature extraction model. We designed a multi-layer extreme learning machine (ML-IELM) with identity mapping in the age estimation stage. We built classifiers and regressors based on ML-IELM for age grouping and final prediction. Our proposed method is evaluated on several prevailing age estimation databases, MORPH-II, Adience Benchmark, and LAP-2016 dataset, in which the performances equal or surpass the state-of-the-art baseline results. In future work, we will do more research at the graph level of computer vision tasks and have better breakthroughs in face recognition.

\section{Acknowledgements}
This work is supported by National Natural Science Foundation of China (Grant No. 61802444); the Changsha Natural Science Foundation (Grant No. kq2202294), the Research Foundation of Education Bureau of Hunan Province of China (Grant No. 20B625, No. 18B196, No. 22B0275); the Research on Local Community Structure Detection Algorithms in Complex Networks (Grant No. 2020YJ009).
\backmatter

\bibliography{sn-bibliography}

\end{document}